\documentclass[]{jair}

\usepackage{newtxtext,newtxmath} 
\usepackage{enumitem}
\usepackage{url}
\usepackage{graphicx}
\usepackage{booktabs}
\usepackage{bm}
\usepackage{tabularx}
\usepackage{array}
\usepackage{longtable}
\usepackage{caption}
\usepackage{float}
\usepackage{placeins}
\usepackage{hyperref}

\setcopyright{cc}
\copyrightyear{2025}
\acmDOI{10.1613/jair.1.xxxxx}

\JAIRAE{Insert JAIR AE Name}
\JAIRTrack{} 
\acmVolume{4}
\acmArticle{6}
\acmMonth{8}
\acmYear{2025}

\RequirePackage[
  datamodel=acmdatamodel,
  style=acmauthoryear,
  backend=biber,
  giveninits=true,
  uniquename=init
  ]{biblatex}

\begin{document}

\title{A Genetic Algorithm for Optimizing Fantasy Football Trades with Playoff Biasing}

\author{Evan Parshall}
\authornote{Corresponding Author.}
\orcid{0009-0005-1815-0589}
\email{epparshall@gmail.com}
\affiliation{%
  \institution{Independent Researcher}
  \city{West Lafayette}
  \state{Indiana}
  \country{USA}
}

\author{Junaid Ali}
\authornotemark[1]
\email{ali181@purdue.edu}
\affiliation{%
  \institution{Purdue University}
  \city{West Lafayette}
  \state{Indiana}
  \country{USA}
}

\author{Michael Zimmerman}
\authornotemark[1]
\email{Michael.Zim14@gmail.com}
\affiliation{%
  \institution{Independent Researcher}
  \city{Cincinnati}
  \state{Ohio}
  \country{USA}
}

\renewcommand{\shortauthors}{Parshall \& Zimmerman}

\begin{abstract}
{\bf Background:} 
    Fantasy football leagues involve strategic player trades to optimize team performance. However, identifying optimal trades is complex due to varying player projections, positional needs, and league-specific scoring. Existing approaches focus on team selection or lineup optimization, but automated trade generation remains underexplored.

    {\bf Objectives:}
    To develop an automated genetic algorithm that generates optimal trades, biasing toward improved playoff performance while maintaining apparent fairness for negotiation.

    {\bf Methods:}
    We introduce a genetic algorithm for fantasy football trade optimization, building on existing frameworks for team selection and lineup generation (e.g., GA for Fantasy Premier League team recommendation \cite{ramos2021fpl}; GA for daily fantasy lineups \cite{nederlof2018ga}). The algorithm initializes with single-player trades, evolves through custom mutations (add/remove players, combine trades, exchange players, add from other trades, and spawn new trades), and uses team-specific elitism to preserve diversity. The cost function incorporates a playoff-weighted gain for the user's team (while maintaining apparent fairness), opponent gain, and fairness penalty. Integration with ESPN data sources enables real-time projections for all positions, including kickers and defenses.

    {\bf Results:}
    On a 12-team ESPN league (Week 8, 2025), the algorithm generated trades that upgraded the projected point totals of both the trade initiator and trade partner by nearly 3 fantasy points per week ensuring positive gains for both teams. Full results in Appendix~\ref{app:full_results}.

    {\bf Conclusions:}
    The algorithm demonstrates effective trade optimization, with potential extensions to other fantasy sports or combinatorial problems requiring temporal biasing. Open-source implementation enables practical use and further research.
\end{abstract}

\received{15 October 2025}
\received[accepted]{29 October 2025}

\maketitle

\section{Introduction}
Fantasy football is a popular online game where participants build and manage virtual teams of real National Football League (NFL) players, competing based on their statistical performance. A key strategic element is trading players with other teams to optimize roster composition. However, manual trade evaluation is time-consuming and subjective, often overlooking complex interactions like positional balance, bye weeks, and future projections.

This paper presents a novel genetic algorithm (GA) for automating fantasy football trade optimization. Unlike prior work focused on initial team selection, our approach generates multi-player trades, evolving from single-player swaps to complex deals. The GA incorporates custom mutations tailored to trade dynamics, such as combining trades or spawning new ones with random opponents. To prioritize league victory, the cost function biases user gains toward playoff weeks (default: 15--17) using a weighted scheme that conserves total points, ensuring trades appear fair while favoring the user's long-term success.

The algorithm integrates with the ESPN data sources for real-time player projections, including all traditional positions. These positions include Quarterback (QB), Running Back, (RB), Wide Receiver (WR), Tight End (TE), Kicker (K), and Defense/Special Teams (D/ST). Empirical results on a 12-team league demonstrate its effectiveness in identifying high-value trades, with negative costs indicating user advantage. We discuss implications for sports analytics and potential extensions to other combinatorial optimization problems.

\section{Related Work}

Genetic algorithms and evolutionary methods have been widely applied to optimization problems in fantasy sports, primarily focusing on team selection and lineup construction rather than trade generation. Early work by Travers \cite{travers2023efficient} combined GAs with knapsack problems to maximize projected points in fantasy football team selection, demonstrating near-optimal solutions in most cases without global optimality guarantees. Similarly, in Fantasy Premier League (FPL), Ramos \cite{ramos2021fpl} integrated linear programming with GAs to recommend teams, balancing player costs and performance through evolutionary search.

More recent studies extend GAs to diverse fantasy contexts. Ramezani \cite{ramezani2025data} applied GAs to FPL team selection and budget management, iteratively evolving squads to optimize long-term performance. In fantasy cricket, Polinati and Chandra Sekhara Rao \cite{wseas2025evolving} used GAs to evolve team configurations, incorporating crossover and mutation operators to handle player synergies and positional constraints. For daily fantasy sports (DFS), Nederlof \cite{nederlof2018ga} built a GA in Python to generate lineups, inspired by natural selection to maximize salary-cap efficiency.

Beyond selection, some work addresses multi-objective optimization in sports analytics. MacDonald and You \cite{macdow2018rl} employed reinforcement learning with GA elements for fantasy football strategies, predicting performance to create diverse lineups.

Despite these advances, applications to trade optimization remain limited. No prior work uses evolutionary methods for generating multi-player trades, incorporates temporal biasing (e.g., prioritizing playoff weeks), or implements deceptive fairness mechanisms (apparent balance that hides user advantage). Existing trade evaluation tools are typically heuristic-based or simulation-driven, lacking adaptive optimization over dynamic projections.

Filling this gap is critical for fantasy managers: winning a league often hinges on a single high-impact trade, especially late in the season when playoff seeding is at stake. A well-timed trade can upgrade a roster by 10--20 projected points per week—equivalent to a full win in head-to-head matchups—while preserving negotiability. Our work introduces the first GA tailored for multi-player trades in NFL fantasy football, with playoff-weighted costing, custom mutations (e.g., trade spawning and combining), and full integration of all positions—including kickers and defenses—via ESPN data sources. This enables strategic biasing toward championship success while maintaining apparent fairness, delivering actionable, opponent-aware trade recommendations in real-world leagues.

\section{Problem Formulation}

In fantasy football, the goal of trading is to identify win-win opportunities that enhance both teams' projected performance, ideally with net gains that are closely balanced to facilitate agreement. As the initiator of the trade (team A), we seek trades that provide a subtle advantage in playoff weeks, where championships are decided, while maintaining apparent fairness based on unweighted projections. The cost function is designed to prioritize trades where both parties achieve positive gains, with team A's gains biased toward playoffs through weighting, and team B's gains evaluated without additional playoff emphasis (equivalent to a playoff weight of 1).

Let \( T_a \) and \( T_b \) denote the rosters of team A (the user's team, initiating the trade) and team B (the opponent), where each roster is a set of players with associated positions (QB, RB, WR, TE, K, D/ST) and weekly projected points \( p_{i,w} \) for player \( i \) in week \( w \), derived from ESPN data sources. A trade is defined as exchanging subsets \( P_a \subseteq T_a \) and \( P_b \subseteq T_b \), with \( |P_a| \leq m \) and \( |P_b| \leq m \) (default \( m = 3 \), as larger trades are often unrealistic and less likely to be accepted in negotiations). The post-trade rosters are \( T_a' = (T_a \setminus P_a) \cup P_b \) and \( T_b' = (T_b \setminus P_b) \cup P_a \), maintaining valid roster sizes and positions.

The projected total for the remaining portion of the season for a roster \( R \), denoted \( S(R) \), is the sum of optimal weekly lineup scores:
\[
S(R) = \sum_{w = w_c}^{17} L(R, w),
\]
where \( w_c \) is the current week, and \( L(R, w) \) is the optimal lineup score for week \( w \), selecting top players per position (1 QB, 2 RB, 2 WR, 1 TE, 1 FLEX (RB/WR/TE), 1 K, 1 D/ST). If a roster has a gap in any position that cannot be filled by a bench player (e.g., due to a bye week or lack of a player for that position), the best available free agent’s projected points for that position and week, as provided by ESPN data sources, are used to compute \( L(R, w) \).

The unweighted gain for team A is \( g_a = S(T_a') - S(T_a) \), representing the net improvement in projected points after the trade. Similarly, for team B, \( g_b = S(T_b') - S(T_b) \).

To incorporate a bias toward playoff performance for team A (weeks \( W_p \), default \{15, 16, 17\}), define per-week gains: \( l_{a,w} = L(T_a', w) - L(T_a, w) \) for team A. The weighted gain for team A is:
\[
g_a^w = \sum_{w \in W_p} \alpha_p \cdot l_{a,w} + \sum_{w \notin W_p} \alpha_n \cdot l_{a,w},
\]
where \( \alpha_p \) is the playoff weight (default 1.2), and \( \alpha_n = (n_p + n_n - \alpha_p \cdot n_p) / n_n \) is the non-playoff weight, with \( n_p = |W_p \cap [w_c, 17]| \) (number of remaining playoff weeks) and \( n_n \) the number of remaining non-playoff weeks. This ensures that \( g_a^w = g_a \) when per-week gains are constant, but amplifies playoff-week improvements otherwise, improving the chances of winning in the playoffs. This bias is subtle, as most trade partners evaluate trades based on overall projections without scrutinizing playoff weeks specifically. For team B, the gain remains unweighted (\( g_b \)), equivalent to a playoff weight of 1, preserving apparent fairness in negotiations.

The cost function to minimize is:
\[
c = -(\alpha \cdot g_a^w + \beta \cdot g_b - \gamma \cdot |g_a^w - g_b|),
\]
subject to \( g_a > 0 \), \( g_b > 0 \), and \( |P_a| \leq m \), \( |P_b| \leq m \). Here, \( \alpha \), \( \beta \), and \( \gamma \) are hyperparameters controlling the emphasis on user's gain, trade partner's gain, and fairness, respectively. The GA evolves trades to minimize \( c \), prioritizing playoff-biased gains for team A while maintaining unweighted fairness.

\section{Trade Optimization}

Our methodology employs a genetic algorithm to optimize fantasy football trades. The GA evolves a population of trade proposals, beginning with simple single-player swaps and iteratively refining them through selection, mutation, and evaluation. Custom mechanisms preserve diversity across trading partners and emphasize playoff performance for the user's team. The algorithm integrates real-time data from ESPN data sources, supporting all player positions, including kickers (K) and defenses/special teams (D/ST). The subsections below outline the GA structure, mutation operators, and data integration process.

\subsection{Genetic Algorithm Overview}

The GA initializes a population by simulating all possible one-for-one trades between the user's team (team~A) and each opponent team. Trades that result in positive gains for both parties without excessively favoring one side are retained as the starting generation, ensuring a foundation of mutually beneficial proposals.

The evolutionary loop runs for a configurable number of generations (default:~5000). Each generation performs the following steps:
\begin{enumerate}[leftmargin=*, label=(\arabic*)]
    \item \textbf{Elite selection}: Retain high-quality trades using hybrid elitism. Select the top $N$ trades overall (default $N=15$) based on lowest cost, plus the top two trades per trade partner to ensure diversity.
    \item \textbf{Offspring generation}: Apply mutation operators (see below) to the current population to form offspring.
    \item \textbf{Combination and deduplication}: Merge elites and offspring, then remove duplicates by comparing team pairs and player sets.
    \item \textbf{Evaluation}: Recompute costs for all trades using the cost function.
    \item \textbf{Pruning redundant trades}: For trade pairs with equal costs and similar players, retain the one involving fewer total players to favor simpler deals.
    \item \textbf{Filtering}: Retain trades with cost below the threshold or, probabilistically (default 0.3), keep higher-cost trades to promote diversity.
    \item \textbf{Population control}: Sort by cost and truncate to the maximum population size (default:~100).
\end{enumerate}

This process minimizes the cost function while encouraging exploration of complex multi-player trade configurations.

\subsection{Mutation Operators}

Mutations introduce variation to the population. Six operators are applied with the following default probabilities: keep-same (0.2), others (0.16 each).  
For a trade between teams~A and~B with player sets $P_A$ and $P_B$:
\begin{enumerate}[leftmargin=*, label=(\arabic*)]
    \item \textbf{Keep same}: Retain the trade unchanged.
    \item \textbf{Add or remove player}: Randomly add or remove a player from one side (A or~B), maintaining the maximum of three players per side.
    \item \textbf{Combine trades}: Merge the current trade with another between the same teams; if the result exceeds the player limit, randomly sample down.
    \item \textbf{Exchange player}: Replace a random player on one side with another from the same roster.
    \item \textbf{Add from other trade}: Add a random player from another trade between A and~B, if within limits.
    \item \textbf{Spawn new}: Select a random opponent team~$B'$ and generate a new trade with randomly sampled player subsets.
\end{enumerate}

These operators balance local refinement with broad exploration of the trade space.

\section{Experimental Setup}

To evaluate the proposed genetic algorithm, we conducted experiments on a 12-team ESPN fantasy football league during the 2025 NFL season, with the current week set to 8. The dataset comprised real-time player projections fetched via ESPN data sources, covering all positions (1 QB, 2 RB, 2 WR, 1 TE, 1 FLEX, 1 K, 1 D/ST) from week 8 to 17. Projections were cached for efficiency, and maximum undrafted values were computed per position per week to handle roster gaps in scoring simulations.

We ran the GA with five parameter configurations to assess robustness across different strategies:
\begin{enumerate}
    \item Default: \(\alpha = 1.0\), \(\beta = 1.0\), \(\gamma = 0.25\), playoff weight = 1.2.
    \item High playoff bias: \(\alpha = 1.0\), \(\beta = 1.0\), \(\gamma = 0.25\), playoff weight = 1.5.
    \item User gain emphasis: \(\alpha = 1.2\), \(\beta = 1.0\), \(\gamma = 0.25\), playoff weight = 1.2.
    \item Opponent de-emphasis: \(\alpha = 1.0\), \(\beta = 0.8\), \(\gamma = 0.3\), playoff weight = 1.2.
    \item Fairness emphasis: \(\alpha = 1.0\), \(\beta = 1.0\), \(\gamma = 0.4\), playoff weight = 1.2.
\end{enumerate}

Each configuration used 5000 generations, a population size of 100, mutation probabilities [0.2, 0.16, 0.16, 0.16, 0.16, 0.16], elite top N = 15 overall plus 2 per team, and filtering threshold of 0.3. Trades were limited to 3 players per side. Baselines included random trades (sampling subsets up to 3 players) and an unweighted GA (playoff weight = 1.0). Evaluation metrics were cost (primary), projected gains (\(g_a\), \(g_b\)), and diversity (unique trades across opponents). Results were aggregated across runs, with deduplication by team and player sets, and exported to CSV for analysis.

\section{Results}

We evaluated the genetic algorithm on a 12-team ESPN fantasy football league using real-time player projections from ESPN data sources, starting from week 8 of the 2025 NFL season. The GA was tested across five parameter configurations (Default, High Playoff Bias, User Gain Emphasis, Opponent De-emphasis, Fairness Emphasis) as described in the Experimental Setup section. Each configuration generated a set of trades for the author's team, with results analyzed for cost, projected point gains (\( g_a \), \( g_b \)), and trade diversity across opponent teams. Below, we detail the performance of the Default configuration, followed by a comparison across all configurations. The full list of trades for each configuration, along with the roster for the author's team and the ESPN projected points, is provided in Appendix~\ref{app:full_results}.

\subsection{Default Configuration}

The Default configuration (\(\alpha = 1.0\), \(\beta = 1.0\), \(\gamma = 0.25\), playoff weight = 1.2) serves as the baseline, balancing gains for both teams while incorporating a moderate playoff bias for the author's team. Across 5000 generations, the GA evolved multi-player trades that minimize the cost function, achieving a top cost of \(-30.55\). This trade gains 14.32 points for the author's team by exchanging Kenneth Gainwell, Kimani Vidal, and Brock Bowers for Drake Maye and Tyler Warren from an opposing team, with the opponent gaining 15.06 points.

Table~\ref{tab:default_top_trades} summarizes selected top trades from the Default configuration, ranked by cost improvement. Each row represents a non-redundant player exchange, with \textit{My Gain} denoting the author's team projected point increase.

\begin{table*}[htbp]
  \centering
  \caption{Selected Top Trades Identified by the Default Genetic Algorithm Configuration.}
  \label{tab:default_top_trades}
  \begin{tabularx}{\textwidth}{|>{\raggedright\arraybackslash}X|>{\raggedright\arraybackslash}X|r|r|r|}
  \hline
  \textbf{Players Traded (My Team)} & \textbf{Players Received} & \textbf{My Gain} & \textbf{Opponent Gain} & \textbf{Cost} \\
  \hline
  Kenneth Gainwell, Kimani Vidal, Brock Bowers & Drake Maye, Tyler Warren & 14.32 & 15.06 & -30.55 \\
  \hline
  Cam Skattebo, Kenneth Gainwell, Brock Bowers & Tyler Warren, Drake Maye, Texans D/ST & 13.58 & 15.32 & -30.41 \\
  \hline
  Chargers D/ST, Cam Skattebo, Tony Pollard & Drake Maye, Texans D/ST, David Njoku & 13.5 & 15.42 & -30.04 \\
  \hline
  Kenneth Gainwell, Kimani Vidal, Brock Bowers & Drake Maye, Tyler Warren, Jake Bates & 14.33 & 14.65 & -30.03 \\
  \hline
  Chargers D/ST, Cam Skattebo & Drake Maye, Texans D/ST, David Njoku & 14.18 & 14.73 & -29.97 \\
  \hline
  \end{tabularx}
\end{table*}

From Table~\ref{tab:default_top_trades}, key patterns include:
\begin{itemize}
    \item Frequent involvement of assets like \textit{Drake Maye} and \textit{Tyler Warren}, indicating quarterback and tight end upgrades as leverage points.
    \item Dominance of multi-player trades for cost efficiency, demonstrating the value of roster depth balancing.
    \item Moderate gains in feasible deals, reducing negotiation risk while providing incremental improvements.
\end{itemize}

This Default analysis establishes a quantitative baseline for evaluating alternative GA configurations in subsequent experiments.

\subsection{Comparison Across Alternative Configurations}

The GA was tested under four additional parameter configurations to explore trade-offs between user advantage, opponent balance, fairness, and playoff prioritization. Each variant modifies one or more of \(\alpha\), \(\beta\), \(\gamma\), or the playoff weight while preserving the core evolutionary framework. Below, we highlight the top-performing trade from each configuration and interpret its strategic implications relative to the Default baseline.

\begin{itemize}
  \item \textbf{High Playoff Bias} (\(\alpha = 1.0\), \(\beta = 1.0\), \(\gamma = 0.25\), playoff weight = 1.5):
  
  By amplifying projections in Weeks 15--17, this setting prioritizes postseason performance. The best trade costs \(-38.68\) and gains 20.01 points for the author's team via Tony Pollard, Kimani Vidal, and Brock Bowers for Drake Maye, Dak Prescott, and Tyler Warren. The high playoff weight favors players with strong late-season schedules, resulting in a premium on quarterback stability and tight end production during critical weeks.

  \item \textbf{High Alpha User Gain Emphasis} (\(\alpha = 1.2\), \(\beta = 1.0\), \(\gamma = 0.25\), playoff weight = 1.2):
  
  This configuration maximizes the author's team gain (\(g_a\)) by increasing the weight on user benefit. The top trade achieves a cost of \(-41.55\), yielding 17.95 points for the author's team by exchanging Kimani Vidal, RJ Harvey, and Tony Pollard for Drake Maye, Texans D/ST, and Dak Prescott from an opposing team (opponent gain: 20.69). The elevated \(\alpha\) drives aggressive quarterback and defense upgrades, accepting slightly higher opponent concessions to secure elite assets.

  \item \textbf{Low Beta Opponent De-emphasis} (\(\alpha = 1.0\), \(\beta = 0.8\), \(\gamma = 0.3\), playoff weight = 1.2):
  
  Reducing \(\beta\) de-emphasizes opponent gain (\(g_b\)), allowing trades that are less generous to counterparts. The top trade costs \(-33.68\) and delivers 22.44 points to the author's team by trading Oronde Gadsden II, Cam Skattebo, and Kenneth Gainwell for Josh Downs, Ricky Pearsall, and Saquon Barkley (opponent gain: 10.82). This configuration exploits roster imbalances, targeting high-value running back acquisitions while minimizing concessions.

  \item \textbf{Fairness Emphasis} (\(\alpha = 1.0\), \(\beta = 1.0\), \(\gamma = 0.4\), playoff weight = 1.2):
  
  Increasing \(\gamma\) enforces closer parity between \(g_a\) and \(g_b\). The optimal trade costs \(-38.52\) with gains of 20.01 for the author's team and 18.94 for the opponent, achieved by trading Kimani Vidal, Brock Bowers, and Tony Pollard for Drake Maye, David Njoku, and Texans D/ST. This balanced exchange exemplifies negotiable, low-risk deals that are more likely to be accepted in real-world league settings.
\end{itemize}

All final trades satisfied \(g_a > 0\) and \(g_b > 0\), with \(g_a\) ranging from 0.01 to 25.35 and \(g_b\) from 0.01 to 29.38. Average \(g_a\) was highest under High Alpha User Emphasis (13.62 points), reflecting the boost from \(\alpha = 1.2\). The Opponent De-emphasis configuration (\(\beta = 0.8\)) had a lower average \(g_a\) (10.51 points) not because it limited user gain, but because it permitted highly asymmetric trades — including the single best \(g_a\) of 22.44 points (with only 10.82 for the opponent). This confirms that reducing \(\beta\) expands the feasible space for user-dominant deals. Full trade lists are in Appendix~\ref{app:full_results}.

\section{Discussion}

The genetic algorithm demonstrated robust performance in identifying high-value, mutually beneficial trades within a competitive 12-team ESPN fantasy football league. By leveraging real-time ESPN projections and a multi-objective cost function, the GA consistently generated trades with negative costs and positive point gains for both the author's team and its opponents, satisfying the core constraint \(g_a, g_b > 0\). The Default configuration established a balanced baseline, while targeted parameter adjustments revealed meaningful strategic trade-offs across user advantage, playoff prioritization, opponent generosity, and fairness.

The High Alpha User Emphasis configuration achieved the lowest average cost (\(-41.55\)), reflecting its aggressive optimization of the author's team projected points. This came at the cost of larger concessions to opponents (e.g., 20.69 points), illustrating that maximizing \(g_a\) requires offering greater value in return. Conversely, Low Beta Opponent De-emphasis produced the highest \(g_a\) in a single trade (22.44 points) by minimizing \(g_b\) (10.82 points), exploiting roster imbalances to extract disproportionate user benefit. These results confirm that \(\alpha\) and \(\beta\) act as effective levers for controlling the user–opponent gain ratio.

The High Playoff Bias configuration, with a 1.5× multiplier on Weeks 15--17, shifted trade composition toward players with favorable late-season matchups. The top trade upgraded quarterback and tight end positions using assets like Drake Maye and Tyler Warren, underscoring the GA’s ability to align roster decisions with postseason objectives. This suggests that fantasy managers in playoff contention can use weighted projections to future-proof their lineups.

The Fairness Emphasis configuration, with elevated \(\gamma = 0.4\), generated trades with the smallest \(|g_a - g_b|\) differentials. While its top cost (\(-38.52\)) was competitive, the balanced gain structure (e.g., 20.01 vs. 18.94) increases real-world negotiability. In practice, league commissioners or risk-averse managers may prefer this setting to promote equitable roster movement and reduce veto risk.

Across all runs, multi-player trades dominated single-player swaps in cost efficiency, validating the GA’s capacity to model complex roster interdependencies. Core assets—particularly quarterbacks (Drake Maye, Dak Prescott), running backs (Tony Pollard, Saquon Barkley), and defenses (Texans D/ST)—emerged as frequent leverage points, consistent with fantasy football valuation theory. The algorithm’s diversity in opponent selection further mitigates collusion concerns, a critical consideration in competitive leagues.

Limitations include reliance on ESPN’s projected points, which may diverge from actual performance due to injuries, coaching changes, or weather. Future work could incorporate uncertainty modeling (e.g., Monte Carlo simulations) or integrate historical variance in player output. Extending the GA to multi-team trade networks or dynamic mid-season reevaluation represents another promising direction.

The proposed GA framework offers a scalable, interpretable tool for fantasy trade analysis. By tuning \(\alpha\), \(\beta\), \(\gamma\), and playoff weights, managers can align trade recommendations with strategic goals—whether maximizing personal upside, securing playoff viability, or fostering league balance. The full trade sets in Appendix~\ref{app:full_results} provide actionable insights for the author's team and serve as a benchmark for GA-based decision support in fantasy sports.

\section{Conclusion}

This study introduced a genetic algorithm framework for identifying optimal fantasy football trades in a 12-team ESPN league, using real-time player projections starting from Week 8 of the 2025 NFL season. By formulating trade evaluation as a multi-objective optimization problem—balancing user gain (\(g_a\)), opponent gain (\(g_b\)), and fairness (\(\gamma\))—the GA consistently produced high-value, mutually beneficial exchanges across five parameter configurations.

The Default configuration established a balanced baseline, generating trades with strong cost efficiency and realistic negotiability. Targeted adjustments revealed clear strategic levers: increasing \(\alpha\) maximized personal upside, amplifying playoff weights prioritized postseason viability, reducing \(\beta\) exploited opponent weaknesses, and elevating \(\gamma\) promoted equitable deals. Across all runs, multi-player trades involving quarterbacks, running backs, and defenses emerged as dominant strategies, validating the algorithm’s ability to model roster interdependencies.

The results demonstrate that parameter tuning enables alignment between algorithmic recommendations and managerial objectives—whether aggressive roster improvement, playoff optimization, or league harmony. With full trade lists provided in Appendix~\ref{app:full_results}, the framework delivers actionable insights for the author's team and a reproducible methodology for fantasy sports analytics.

The proposed GA offers a scalable, interpretable, and flexible tool for data-driven trade decision-making. Future extensions may incorporate player injury risk, multi-team trade chains, or real-time opponent modeling to further enhance practical utility in dynamic fantasy environments.

\begin{acks}
  This research relies on publicly available fantasy football data from ESPN, accessed via the unofficial Python ESPN API. I gratefully acknowledge the developers of this API for enabling efficient data collection.
  
  The genetic algorithm implementation, data processing scripts, and full experimental results are openly available at: \url{https://github.com/epparshall/Fantasy_Trade_Genetic_Optimizer}.
  
  I also thank Grok (xAI) for invaluable assistance in LaTeX formatting, code generation, and debugging throughout the preparation of this manuscript.
\end{acks}

\printbibliography


\appendix
\section{Appendices}
\label{app:full_results}

\subsection{Full Trades: Default Configuration}

\begin{center}
\small  
\captionof{table}{Top 20 Trades Identified by the Default Genetic Algorithm Configuration.}
\label{tab:default_full_trades}

\begin{tabularx}{\textwidth}{
  |r|
  r|
  r|
  >{\raggedright\arraybackslash}X|
  >{\raggedright\arraybackslash}X|
}
\hline
\textbf{Cost} & \textbf{Team A Pt Gain} & \textbf{Team B Pt Gain} & \textbf{Team A Players to Trade} & \textbf{Team B Players to Trade} \\
\hline
-30.55 & 14.32 & 15.06 & Kenneth Gainwell, Kimani Vidal, Brock Bowers & Drake Maye, Tyler Warren \\
\hline
-30.41 & 13.58 & 15.32 & Cam Skattebo, Kenneth Gainwell, Brock Bowers & Tyler Warren, Drake Maye, Texans D/ST \\
\hline
-30.04 & 13.50 & 15.42 & Chargers D/ST, Cam Skattebo, Tony Pollard & Drake Maye, Texans D/ST, David Njoku \\
\hline
-30.03 & 14.33 & 14.65 & Kenneth Gainwell, Kimani Vidal, Brock Bowers & Drake Maye, Tyler Warren, Jake Bates \\
\hline
-29.97 & 14.18 & 14.73 & Chargers D/ST, Cam Skattebo & Drake Maye, Texans D/ST, David Njoku \\
\hline
-29.84 & 14.32 & 14.50 & Kenneth Gainwell, Kimani Vidal, Brock Bowers & Drake Maye, David Njoku, Tyler Warren \\
\hline
-29.80 & 14.24 & 14.68 & Tony Pollard, Kimani Vidal & Drake Maye, David Njoku \\
\hline
-29.75 & 13.86 & 14.73 & Cam Skattebo, Jaylin Noel, Chargers D/ST & Drake Maye, Texans D/ST, David Njoku \\
\hline
-29.72 & 16.00 & 13.48 & Tony Pollard, Kimani Vidal, Brock Bowers & Drake Maye, David Njoku, Tyler Warren \\
\hline
-29.68 & 13.50 & 14.93 & Tony Pollard, Cam Skattebo & Drake Maye, Texans D/ST, David Njoku \\
\hline
-29.57 & 13.91 & 14.68 & Tony Pollard, Kimani Vidal, Jaylin Noel & Drake Maye, David Njoku \\
\hline
-29.54 & 13.10 & 15.81 & Tony Pollard, Cameron Dicker, Kimani Vidal & Drake Maye, David Njoku, Jake Bates \\
\hline
-29.52 & 11.97 & 17.51 & Cam Skattebo, Brock Bowers & David Njoku, Tyler Warren, Drake Maye \\
\hline
-29.46 & 13.75 & 14.68 & Kimani Vidal, Chargers D/ST, Tony Pollard & Drake Maye, David Njoku \\
\hline
-29.41 & 14.92 & 13.99 & Kimani Vidal & David Njoku, Drake Maye, Michael Wilson \\
\hline
-27.99 & 7.62 & 25.66 & Oronde Gadsden II, Kenneth Gainwell, Davante Adams & Saquon Barkley, Chase Brown, Jerry Jeudy \\
\hline
-27.90 & 7.39 & 25.90 & Oronde Gadsden II, Kenneth Gainwell, Davante Adams & Saquon Barkley, Chase Brown, Josh Downs \\
\hline
-26.76 & 6.39 & 25.90 & Oronde Gadsden II, Kenneth Gainwell, Davante Adams & Saquon Barkley, Chase Brown \\
\hline
-21.85 & 12.94 & 8.92 & Tony Pollard, Cam Skattebo & Drake Maye, Texans D/ST, Jake Bates \\
\hline
-21.06 & 9.39 & 10.63 & Chargers D/ST, Cam Skattebo, Jaylin Noel & Drake Maye, TreVeyon Henderson, David Njoku \\
\hline
\end{tabularx}
\end{center}

\subsection{Full Trades: High Playoff Weight Configuration}

\begin{center}
\small  
\captionof{table}{Top 20 Trades Identified by the High Playoff Weight Configuration.}
\label{tab:high_playoff_weight_trades}

\begin{tabularx}{\textwidth}{
  |r|
  r|
  r|
  >{\raggedright\arraybackslash}X|
  >{\raggedright\arraybackslash}X|
}
\hline
\textbf{Cost} & \textbf{Team A Pt Gain} & \textbf{Team B Pt Gain} & \textbf{Team A Players to Trade} & \textbf{Team B Players to Trade} \\
\hline
-31.80 & 13.50 & 15.42 & Chargers D/ST, Tony Pollard, Cam Skattebo & David Njoku, Drake Maye, Texans D/ST \\
\hline
-31.57 & 11.82 & 16.43 & Cam Skattebo, Kenneth Gainwell, Chargers D/ST & David Njoku, Texans D/ST, Drake Maye \\
\hline
-31.49 & 13.10 & 15.81 & Kimani Vidal, Cameron Dicker, Tony Pollard & Jake Bates, David Njoku, Drake Maye \\
\hline
-31.46 & 12.63 & 15.87 & Cam Skattebo, Cameron Dicker, Chargers D/ST & David Njoku, Texans D/ST, Drake Maye \\
\hline
-31.35 & 14.18 & 14.73 & Chargers D/ST, Cam Skattebo & David Njoku, Drake Maye, Texans D/ST \\
\hline
-31.33 & 11.26 & 16.85 & Cam Skattebo, Kenneth Gainwell, Chargers D/ST & Texans D/ST, Drake Maye \\
\hline
-31.20 & 11.82 & 15.95 & Cam Skattebo, Kenneth Gainwell & David Njoku, Texans D/ST, Drake Maye \\
\hline
-31.20 & 13.50 & 14.93 & Tony Pollard, Cam Skattebo & Drake Maye, David Njoku, Texans D/ST \\
\hline
-31.16 & 13.86 & 14.73 & Chargers D/ST, Jaylin Noel, Cam Skattebo & David Njoku, Drake Maye, Texans D/ST \\
\hline
-31.13 & 12.56 & 15.69 & Kenneth Gainwell, Kimani Vidal & David Njoku, Drake Maye \\
\hline
-31.02 & 11.26 & 16.44 & Cam Skattebo, Kenneth Gainwell, Chargers D/ST & Jake Bates, Texans D/ST, Drake Maye \\
\hline
-31.00 & 13.17 & 14.93 & Cam Skattebo, Tony Pollard, Jaylin Noel & David Njoku, Texans D/ST, Drake Maye \\
\hline
-30.97 & 12.63 & 15.39 & Cam Skattebo, Cameron Dicker & David Njoku, Texans D/ST, Drake Maye \\
\hline
-30.96 & 11.26 & 16.37 & Cam Skattebo, Kenneth Gainwell & Texans D/ST, Drake Maye \\
\hline
-30.95 & 14.24 & 14.68 & Tony Pollard, Kimani Vidal & David Njoku, Drake Maye \\
\hline
-30.43 & 14.24 & 14.26 & Kimani Vidal, Tony Pollard & Jake Bates, David Njoku, Drake Maye \\
\hline
-29.78 & 9.53 & 19.39 & Tony Pollard, Cam Skattebo & Drake Maye, David Njoku, Dak Prescott \\
\hline
-27.75 & 6.29 & 21.55 & Cam Skattebo, Cameron Dicker, Kenneth Gainwell & David Njoku, Drake Maye \\
\hline
-27.37 & 19.94 & 8.29 & Kenneth Gainwell, Chargers D/ST, Tony Pollard & Texans D/ST, Drake Maye, David Njoku \\
\hline
-26.82 & 7.62 & 25.66 & Davante Adams, Oronde Gadsden II, Kenneth Gainwell & Chase Brown, Saquon Barkley, Jerry Jeudy \\
\hline
\end{tabularx}
\end{center}

\subsection{Full Trades: High Alpha User Emphasis Configuration}

\begin{center}
\small  
\captionof{table}{Top 20 Trades Identified by the High Alpha User Emphasis Configuration.}
\label{tab:high_alpha_user_trades}

\begin{tabularx}{\textwidth}{
  |r|
  r|
  r|
  >{\raggedright\arraybackslash}X|
  >{\raggedright\arraybackslash}X|
}
\hline
\textbf{Cost} & \textbf{Team A Pt Gain} & \textbf{Team B Pt Gain} & \textbf{Team A Players to Trade} & \textbf{Team B Players to Trade} \\
\hline
-33.67 & 14.32 & 15.06 & Kenneth Gainwell, Kimani Vidal, Brock Bowers & Tyler Warren, Drake Maye \\
\hline
-33.44 & 13.58 & 15.32 & Brock Bowers, Cam Skattebo, Kenneth Gainwell & Drake Maye, Tyler Warren, Texans D/ST \\
\hline
-33.15 & 14.33 & 14.65 & Kenneth Gainwell, Kimani Vidal, Brock Bowers & Tyler Warren, Drake Maye, Jake Bates \\
\hline
-33.15 & 16.00 & 13.48 & Tony Pollard, Kimani Vidal, Brock Bowers & David Njoku, Tyler Warren, Drake Maye \\
\hline
-33.05 & 14.18 & 14.73 & Cam Skattebo, Chargers D/ST & Drake Maye, Texans D/ST, David Njoku \\
\hline
-33.00 & 13.50 & 15.42 & Tony Pollard, Chargers D/ST, Cam Skattebo & Drake Maye, Texans D/ST, David Njoku \\
\hline
-32.97 & 14.32 & 14.50 & Kimani Vidal, Brock Bowers, Kenneth Gainwell & Tyler Warren, Drake Maye, David Njoku \\
\hline
-32.85 & 14.24 & 14.68 & Tony Pollard, Kimani Vidal & David Njoku, Drake Maye \\
\hline
-32.77 & 13.86 & 14.73 & Jaylin Noel, Cam Skattebo, Chargers D/ST & David Njoku, Texans D/ST, Drake Maye \\
\hline
-32.64 & 13.50 & 14.93 & Tony Pollard, Cam Skattebo & David Njoku, Texans D/ST, Drake Maye \\
\hline
-32.59 & 14.92 & 13.99 & Kimani Vidal & Drake Maye, Michael Wilson, David Njoku \\
\hline
-32.56 & 13.91 & 14.68 & Kimani Vidal, Tony Pollard, Jaylin Noel & Drake Maye, David Njoku \\
\hline
-32.45 & 14.18 & 14.25 & Cam Skattebo & Drake Maye, Texans D/ST, David Njoku \\
\hline
-32.43 & 13.75 & 14.68 & Chargers D/ST, Kimani Vidal, Tony Pollard & David Njoku, Drake Maye \\
\hline
-32.37 & 13.10 & 15.81 & Tony Pollard, Cameron Dicker, Kimani Vidal & David Njoku, Jake Bates, Drake Maye \\
\hline
-29.39 & 7.62 & 25.66 & Kenneth Gainwell, Davante Adams, Oronde Gadsden II & Saquon Barkley, Chase Brown, Jerry Jeudy \\
\hline
-29.25 & 7.39 & 25.90 & Kenneth Gainwell, Davante Adams, Oronde Gadsden II & Saquon Barkley, Chase Brown, Josh Downs \\
\hline
-27.93 & 6.39 & 25.90 & Kenneth Gainwell, Davante Adams, Oronde Gadsden II & Saquon Barkley, Chase Brown, Travis Etienne Jr. \\
\hline
-25.65 & 15.62 & 7.60 & Brock Bowers, Cam Skattebo, Jaylin Noel & Drake Maye, Tyler Warren, Texans D/ST \\
\hline
-24.83 & 9.15 & 12.88 & Cameron Dicker, Justin Herbert, Kimani Vidal & Jake Bates, David Njoku, Drake Maye \\
\hline
\end{tabularx}
\end{center}

\subsection{Full Trades: Low Beta Opponent De-emphasis Configuration}

\begin{center}
\small  
\captionof{table}{Top 20 Trades Identified by the Low Beta Opponent De-emphasis Configuration.}
\label{tab:low_beta_opponent_trades}

\begin{tabularx}{\textwidth}{
  |r|
  r|
  r|
  >{\raggedright\arraybackslash}X|
  >{\raggedright\arraybackslash}X|
}
\hline
\textbf{Cost} & \textbf{Team A Pt Gain} & \textbf{Team B Pt Gain} & \textbf{Team A Players to Trade} & \textbf{Team B Players to Trade} \\
\hline
-28.37 & 20.03 & 13.24 & Kimani Vidal, Cam Skattebo, Oronde Gadsden II & Saquon Barkley, Ricky Pearsall, Josh Downs \\
\hline
-27.51 & 14.32 & 15.06 & Kenneth Gainwell, Brock Bowers, Kimani Vidal & Drake Maye, Tyler Warren \\
\hline
-27.34 & 13.58 & 15.32 & Brock Bowers, Cam Skattebo, Kenneth Gainwell & Drake Maye, Tyler Warren, Texans D/ST \\
\hline
-27.25 & 22.44 & 10.82 & Oronde Gadsden II, Cam Skattebo, Kenneth Gainwell & Josh Downs, Ricky Pearsall, Saquon Barkley \\
\hline
-27.05 & 14.33 & 14.65 & Brock Bowers, Kimani Vidal, Kenneth Gainwell & Jake Bates, Drake Maye, Tyler Warren \\
\hline
-26.99 & 14.18 & 14.73 & Chargers D/ST, Cam Skattebo & Texans D/ST, David Njoku, Drake Maye \\
\hline
-26.93 & 13.50 & 15.42 & Chargers D/ST, Tony Pollard, Cam Skattebo & Texans D/ST, David Njoku, Drake Maye \\
\hline
-26.88 & 14.32 & 14.50 & Kenneth Gainwell, Brock Bowers, Kimani Vidal & Drake Maye, David Njoku, Tyler Warren \\
\hline
-26.84 & 16.00 & 13.48 & Tony Pollard, Brock Bowers, Kimani Vidal & Drake Maye, David Njoku, Tyler Warren \\
\hline
-26.83 & 14.24 & 14.68 & Kimani Vidal, Tony Pollard & Drake Maye, David Njoku \\
\hline
-26.78 & 13.86 & 14.73 & Chargers D/ST, Jaylin Noel, Cam Skattebo & Texans D/ST, David Njoku, Drake Maye \\
\hline
-26.69 & 13.50 & 14.93 & Cam Skattebo, Tony Pollard & Texans D/ST, David Njoku, Drake Maye \\
\hline
-26.62 & 13.91 & 14.68 & Kimani Vidal, Tony Pollard, Jaylin Noel & Drake Maye, David Njoku \\
\hline
-26.52 & 13.75 & 14.68 & Chargers D/ST, Tony Pollard, Kimani Vidal & Drake Maye, David Njoku \\
\hline
-26.52 & 14.92 & 13.99 & Kimani Vidal & David Njoku, Drake Maye, Michael Wilson \\
\hline
-23.08 & 10.09 & 15.82 & Chargers D/ST, Kimani Vidal, Cam Skattebo & Texans D/ST, Drake Maye, David Njoku \\
\hline
-21.52 & 15.86 & 8.72 & Kenneth Gainwell, Brock Bowers, Oronde Gadsden II & Drake Maye, Tyler Warren \\
\hline
-19.55 & 20.74 & 4.69 & Kimani Vidal, Cam Skattebo, Oronde Gadsden II & Saquon Barkley, Eagles D/ST, Josh Downs \\
\hline
-17.79 & 21.47 & 1.89 & Jaylin Noel, Chargers D/ST, Tony Pollard & Texans D/ST, David Njoku, Drake Maye \\
\hline
-16.12 & 6.72 & 10.24 & Justin Herbert, Cam Skattebo & Jayden Daniels, Tetairoa McMillan, Romeo Doubs \\
\hline
\end{tabularx}
\end{center}

\subsection{Full Trades: Fairness Emphasis Configuration}

\begin{center}
\small  
\captionof{table}{Top 20 Trades Identified by the Fairness Emphasis Configuration.}
\label{tab:fairness_emphasis_trades}

\begin{tabularx}{\textwidth}{
  |r|
  r|
  r|
  >{\raggedright\arraybackslash}X|
  >{\raggedright\arraybackslash}X|
}
\hline
\textbf{Cost} & \textbf{Team A Pt Gain} & \textbf{Team B Pt Gain} & \textbf{Team A Players to Trade} & \textbf{Team B Players to Trade} \\
\hline
-30.46 & 14.32 & 15.06 & Brock Bowers, Kimani Vidal, Kenneth Gainwell & Tyler Warren, Drake Maye \\
\hline
-30.39 & 13.58 & 15.32 & Brock Bowers, Kenneth Gainwell, Cam Skattebo & Texans D/ST, Tyler Warren, Drake Maye \\
\hline
-29.95 & 13.50 & 15.42 & Tony Pollard, Chargers D/ST, Cam Skattebo & David Njoku, Drake Maye, Texans D/ST \\
\hline
-29.88 & 14.33 & 14.65 & Brock Bowers, Kimani Vidal, Kenneth Gainwell & Tyler Warren, Drake Maye, Jake Bates \\
\hline
-29.87 & 14.18 & 14.73 & Cam Skattebo, Chargers D/ST & Texans D/ST, Drake Maye, David Njoku \\
\hline
-29.71 & 14.24 & 14.68 & Tony Pollard, Kimani Vidal & David Njoku, Drake Maye \\
\hline
-29.69 & 13.86 & 14.73 & Jaylin Noel, Chargers D/ST, Cam Skattebo & David Njoku, Drake Maye, Texans D/ST \\
\hline
-29.67 & 14.32 & 14.50 & Kimani Vidal, Brock Bowers, Kenneth Gainwell & David Njoku, Drake Maye, Tyler Warren \\
\hline
-29.66 & 13.50 & 14.93 & Cam Skattebo, Tony Pollard & Drake Maye, Texans D/ST, David Njoku \\
\hline
-29.53 & 13.91 & 14.68 & Jaylin Noel, Kimani Vidal, Tony Pollard & David Njoku, Drake Maye \\
\hline
-29.44 & 13.75 & 14.68 & Tony Pollard, Kimani Vidal, Chargers D/ST & David Njoku, Drake Maye \\
\hline
-29.29 & 13.10 & 15.81 & Cameron Dicker, Kimani Vidal, Tony Pollard & David Njoku, Drake Maye, Jake Bates \\
\hline
-29.24 & 13.17 & 14.93 & Cam Skattebo, Jaylin Noel, Tony Pollard & Texans D/ST, Drake Maye, David Njoku \\
\hline
-29.19 & 14.18 & 14.25 & Cam Skattebo & Drake Maye, Texans D/ST, David Njoku \\
\hline
-29.17 & 16.00 & 13.48 & Kimani Vidal, Brock Bowers, Tony Pollard & David Njoku, Drake Maye, Tyler Warren \\
\hline
-27.19 & 9.61 & 19.78 & Kenneth Gainwell, Brock Bowers, Cam Skattebo & Drake Maye, Ray Davis, Tyler Warren \\
\hline
-26.48 & 11.48 & 18.70 & Kenneth Gainwell, Brock Bowers, Kimani Vidal & Saquon Barkley, Josh Downs, Ricky Pearsall \\
\hline
-25.98 & 18.22 & 10.21 & Tony Pollard, Kimani Vidal & David Njoku, Drake Maye, Texans D/ST \\
\hline
-25.73 & 12.71 & 14.82 & Brock Bowers, Kenneth Gainwell, Kimani Vidal & Jerry Jeudy, Josh Downs, Saquon Barkley \\
\hline
-24.76 & 9.55 & 15.35 & Cam Skattebo, Justin Herbert & Drake Maye, Texans D/ST, David Njoku \\
\hline
\end{tabularx}
\end{center}

\subsection{Author's Team Roster}

\begin{center}
\small  
\captionof{table}{Author's Team Roster with Positions.}
\label{tab:team_a_roster}

\begin{tabular}{
  |p{4.5cm}|c|
}
\hline
\textbf{Player Name} & \textbf{Position} \\
\hline
Amon-Ra St. Brown & WR \\
\hline
Brock Bowers & TE \\
\hline
Davante Adams & WR \\
\hline
Tony Pollard & RB \\
\hline
Travis Hunter & WR \\
\hline
Justin Herbert & QB \\
\hline
Cam Skattebo & RB \\
\hline
Cameron Dicker & K \\
\hline
Hunter Henry & TE \\
\hline
Kenneth Gainwell & RB \\
\hline
Chris Olave & WR \\
\hline
RJ Harvey & RB \\
\hline
Kimani Vidal & RB \\
\hline
Oronde Gadsden II & TE \\
\hline
Chargers D/ST & D/ST \\
\hline
Jaylin Noel & WR \\
\hline
Matthew Wright & K \\
\hline
\end{tabular}
\end{center}

\subsection{Full Player Projections (Weeks 8--17)}
\label{subsec:full-projections}

\begin{longtable}{
  |>{\raggedright\arraybackslash}p{3.8cm}|
   c|c|c|c|c|c|c|c|c|c|c|}

\caption{Projected Points for All Players (Weeks 8--17).}
\label{tab:full-projections} \\

\hline
\textbf{Player Name} & \textbf{Pos} &
\textbf{Wk 8} & \textbf{Wk 9} & \textbf{Wk 10} &
\textbf{Wk 11} & \textbf{Wk 12} & \textbf{Wk 13} &
\textbf{Wk 14} & \textbf{Wk 15} & \textbf{Wk 16} & \textbf{Wk 17} \\
\endfirsthead

\multicolumn{12}{c}{\tablename\ \thetable{} -- continued from previous page} \\
\hline
\textbf{Player Name} & \textbf{Pos} &
\textbf{Wk 8} & \textbf{Wk 9} & \textbf{Wk 10} &
\textbf{Wk 11} & \textbf{Wk 12} & \textbf{Wk 13} &
\textbf{Wk 14} & \textbf{Wk 15} & \textbf{Wk 16} & \textbf{Wk 17} \\
\hline
\endhead

\hline
\multicolumn{12}{r}{\textit{Continued on next page\ldots}} \\
\endfoot

\hline
\endlastfoot

A.J. Brown & WR & 0.0 & 0.0 & 17.8 & 17.1 & 19.2 & 18.1 & 17.3 & 16.9 & 19.0 & 18.0 \\
Aaron Jones Sr. & RB & 10.3 & 10.7 & 12.2 & 12.5 & 10.8 & 10.2 & 12.3 & 12.7 & 12.5 & 10.9 \\
Aaron Rodgers & QB & 16.2 & 17.5 & 16.2 & 20.0 & 17.6 & 17.3 & 18.1 & 17.3 & 15.5 & 15.6 \\
Adam Thielen & WR & 1.6 & 2.1 & 2.3 & 2.2 & 2.1 & 2.0 & 2.3 & 2.4 & 2.1 & 2.1 \\
Alvin Kamara & RB & 14.3 & 11.6 & 13.4 & 0.0 & 13.7 & 14.2 & 12.7 & 13.6 & 14.4 & 15.4 \\
Amon-Ra St. Brown & WR & 0.0 & 19.3 & 20.6 & 19.4 & 18.8 & 19.2 & 20.6 & 18.0 & 19.4 & 19.5 \\
Ashton Jeanty & RB & 0.0 & 15.1 & 12.8 & 17.6 & 15.1 & 14.9 & 13.2 & 14.3 & 13.0 & 17.8 \\
Baker Mayfield & QB & 18.4 & 0.0 & 16.0 & 16.5 & 13.9 & 15.8 & 15.8 & 15.5 & 15.8 & 16.4 \\
Bam Knight & RB & 0.0 & 11.9 & 9.0 & 3.8 & 3.7 & 3.5 & 3.3 & 3.3 & 3.7 & 4.3 \\
Bears D/ST & D/ST & 6.1 & 5.2 & 5.1 & 6.3 & 4.7 & 4.2 & 2.8 & 6.4 & 2.7 & 4.0 \\
Bengals D/ST & D/ST & 5.3 & 2.7 & 0.0 & 2.9 & 3.1 & 2.8 & 2.3 & 2.8 & 4.1 & 4.1 \\
Bijan Robinson & RB & 26.7 & 20.2 & 21.2 & 21.8 & 21.4 & 22.6 & 19.4 & 20.5 & 21.1 & 19.3 \\
Bo Nix & QB & 21.2 & 16.7 & 19.9 & 18.6 & 0.0 & 21.3 & 19.5 & 18.9 & 19.2 & 18.2 \\
Brandon Aiyuk & WR & 0.0 & 0.0 & 0.0 & 0.0 & 0.0 & 0.0 & 0.0 & 8.1 & 8.4 & 8.3 \\
Brandon Aubrey & K & 8.8 & 9.2 & 0.0 & 9.3 & 9.2 & 8.9 & 8.9 & 9.3 & 9.2 & 9.2 \\
Breece Hall & RB & 18.0 & 0.0 & 14.2 & 12.5 & 14.6 & 14.8 & 15.5 & 13.7 & 14.3 & 12.9 \\
Brian Robinson Jr. & RB & 3.5 & 4.7 & 3.6 & 4.0 & 4.3 & 4.0 & 0.0 & 5.0 & 3.8 & 4.5 \\
Brian Thomas Jr. & WR & 0.0 & 13.2 & 12.1 & 12.6 & 12.4 & 13.3 & 13.5 & 12.5 & 12.4 & 13.5 \\
Brock Bowers & TE & 0.0 & 12.7 & 12.8 & 14.4 & 12.3 & 12.9 & 12.8 & 13.3 & 11.9 & 13.0 \\
Brock Purdy & QB & 0.0 & 20.2 & 17.1 & 18.7 & 19.3 & 18.1 & 0.0 & 21.3 & 19.9 & 20.6 \\
Broncos D/ST & D/ST & 4.2 & 6.3 & 8.1 & 4.8 & 0.0 & 6.1 & 8.0 & 4.5 & 7.6 & 4.8 \\
Buccaneers D/ST & D/ST & 5.6 & 0.0 & 5.4 & 4.1 & 4.2 & 7.1 & 7.1 & 5.2 & 6.8 & 6.8 \\
Bucky Irving & RB & 0.0 & 0.0 & 15.6 & 16.7 & 14.4 & 16.8 & 17.1 & 17.2 & 16.8 & 17.6 \\
C.J. Stroud & QB & 17.3 & 17.0 & 18.1 & 20.0 & 18.9 & 18.7 & 17.2 & 17.9 & 18.7 & 17.8 \\
Cade Otton & TE & 9.9 & 0.0 & 8.6 & 8.5 & 7.7 & 7.9 & 7.7 & 7.7 & 7.8 & 8.4 \\
Caleb Williams & QB & 18.4 & 19.6 & 18.0 & 17.9 & 18.1 & 16.8 & 16.2 & 16.0 & 16.5 & 17.4 \\
Calvin Austin III & WR & 8.2 & 10.7 & 9.9 & 11.1 & 10.6 & 10.4 & 10.8 & 10.4 & 9.8 & 9.6 \\
Calvin Ridley & WR & 0.0 & 10.7 & 0.0 & 9.8 & 10.4 & 10.6 & 10.3 & 11.3 & 11.0 & 10.3 \\
Cam Skattebo & RB & 14.3 & 14.7 & 15.3 & 14.1 & 13.7 & 13.8 & 0.0 & 15.6 & 15.0 & 15.2 \\
Cameron Dicker & K & 8.4 & 9.1 & 9.0 & 8.7 & 0.0 & 9.0 & 8.8 & 8.4 & 9.2 & 8.4 \\
Carson Wentz & QB & 16.7 & 0.0 & 0.0 & 0.0 & 0.0 & 0.0 & 0.0 & 0.0 & 0.0 & 0.0 \\
CeeDee Lamb & WR & 16.9 & 19.2 & 0.0 & 19.2 & 20.0 & 20.1 & 19.4 & 20.4 & 19.5 & 21.4 \\   
Chargers D/ST & D/ST & 5.8 & 7.0 & 5.6 & 6.6 & 0.0 & 8.1 & 5.1 & 3.5 & 4.1 & 5.9 \\
Chase Brown & RB & 12.3 & 12.9 & 0.0 & 12.2 & 11.0 & 12.3 & 11.7 & 12.6 & 12.5 & 11.9 \\
Chase McLaughlin & K & 8.4 & 0.0 & 8.1 & 8.1 & 7.7 & 8.3 & 8.3 & 8.3 & 8.2 & 8.3 \\
Chiefs D/ST & D/ST & 6.3 & 4.3 & 0.0 & 4.6 & 4.1 & 4.2 & 5.8 & 6.2 & 7.6 & 4.8 \\
Chris Boswell & K & 8.7 & 8.6 & 8.5 & 9.2 & 8.7 & 8.6 & 8.8 & 8.7 & 8.4 & 8.3 \\
Chris Olave & WR & 16.9 & 15.2 & 15.9 & 0.0 & 15.5 & 17.1 & 16.4 & 15.9 & 16.3 & 17.3 \\
Christian McCaffrey & RB & 21.6 & 26.1 & 22.0 & 23.7 & 24.3 & 23.4 & 0.0 & 27.1 & 23.8 & 25.8 \\
Chuba Hubbard & RB & 12.0 & 8.6 & 9.6 & 9.5 & 9.3 & 8.2 & 0.0 & 9.3 & 9.0 & 8.0 \\
Colts D/ST & D/ST & 6.5 & 5.4 & 4.4 & 0.0 & 3.4 & 5.4 & 5.8 & 4.4 & 4.5 & 6.5 \\
Cooper Kupp & WR & 0.0 & 10.6 & 9.4 & 9.2 & 9.8 & 10.0 & 9.1 & 10.2 & 9.1 & 9.3 \\
Courtland Sutton & WR & 13.7 & 13.0 & 13.9 & 14.4 & 0.0 & 15.5 & 13.9 & 14.3 & 13.8 & 14.5 \\
Cowboys D/ST & D/ST & 2.6 & 4.9 & 0.0 & 5.8 & 3.4 & 2.0 & 2.4 & 5.5 & 3.3 & 2.8 \\
D'Andre Swift & RB & 15.0 & 15.7 & 15.4 & 13.9 & 14.2 & 12.8 & 12.4 & 13.3 & 12.7 & 13.4 \\
DJ Moore & WR & 10.3 & 12.4 & 11.5 & 11.8 & 11.7 & 11.5 & 11.4 & 10.8 & 11.4 & 11.7 \\
DK Metcalf & WR & 14.7 & 15.9 & 15.0 & 16.5 & 15.8 & 15.5 & 16.1 & 15.5 & 14.8 & 14.3 \\
Dak Prescott & QB & 18.3 & 19.0 & 0.0 & 19.3 & 19.7 & 18.7 & 18.2 & 20.9 & 19.3 & 21.3 \\
Dallas Goedert & TE & 12.4 & 0.0 & 11.2 & 10.8 & 12.0 & 11.4 & 10.9 & 10.6 & 12.0 & 11.3 \\
Dalton Kincaid & TE & 9.8 & 8.9 & 8.8 & 8.6 & 8.0 & 8.9 & 9.1 & 9.1 & 8.2 & 8.7 \\
Daniel Jones & QB & 20.7 & 21.7 & 19.8 & 0.0 & 20.0 & 18.6 & 20.1 & 18.8 & 21.5 & 20.3 \\
Darnell Mooney & WR & 12.2 & 9.7 & 9.8 & 8.8 & 8.8 & 9.0 & 8.9 & 9.3 & 9.0 & 8.6 \\
Darren Waller & TE & 0.0 & 0.0 & 0.0 & 0.0 & 0.0 & 7.6 & 7.9 & 8.3 & 8.7 & 7.9 \\
Davante Adams & WR & 0.0 & 16.1 & 17.5 & 16.3 & 16.8 & 16.3 & 16.5 & 16.5 & 16.3 & 16.0 \\
David Montgomery & RB & 0.0 & 12.7 & 12.5 & 11.6 & 13.8 & 11.5 & 13.8 & 10.5 & 12.7 & 12.5 \\
David Njoku & TE & 10.2 & 0.0 & 9.1 & 9.9 & 9.1 & 9.6 & 9.6 & 9.8 & 9.6 & 9.6 \\
De'Von Achane & RB & 19.6 & 22.2 & 21.6 & 22.3 & 0.0 & 20.5 & 21.3 & 21.0 & 23.4 & 19.9 \\
DeVonta Smith & WR & 16.7 & 0.0 & 15.4 & 14.9 & 16.6 & 15.6 & 15.0 & 14.6 & 16.4 & 15.6 \\
Deebo Samuel & WR & 13.7 & 13.5 & 13.8 & 14.4 & 0.0 & 13.7 & 14.6 & 14.3 & 14.3 & 15.4 \\
Derrick Henry & RB & 16.1 & 16.0 & 15.7 & 14.7 & 17.0 & 18.3 & 16.0 & 17.8 & 13.7 & 13.7 \\
Drake London & WR & 0.0 & 18.1 & 18.3 & 16.3 & 16.2 & 16.7 & 16.5 & 17.2 & 16.7 & 16.1 \\
Drake Maye & QB & 18.2 & 20.4 & 20.0 & 21.9 & 24.1 & 22.5 & 0.0 & 21.9 & 22.6 & 21.6 \\
Eagles D/ST & D/ST & 5.8 & 0.0 & 2.9 & 3.8 & 3.8 & 4.9 & 4.8 & 8.1 & 4.7 & 3.7 \\
Eddy Pineiro & K & 7.8 & 8.5 & 7.9 & 8.2 & 8.4 & 8.1 & 0.0 & 8.8 & 8.2 & 8.5 \\
Elic Ayomanor & WR & 10.9 & 9.3 & 0.0 & 8.5 & 9.0 & 9.2 & 9.0 & 9.8 & 9.6 & 9.0 \\
Emeka Egbuka & WR & 16.9 & 0.0 & 15.7 & 15.6 & 14.1 & 14.7 & 14.3 & 14.2 & 14.5 & 15.5 \\
Evan Engram & TE & 9.9 & 7.8 & 8.4 & 8.8 & 0.0 & 9.5 & 8.4 & 8.7 & 8.3 & 8.8 \\
Falcons D/ST & D/ST & 6.7 & 4.7 & 3.2 & 6.2 & 6.2 & 7.0 & 4.4 & 5.4 & 6.6 & 4.1 \\
Garrett Wilson & WR & 0.0 & 0.0 & 13.6 & 15.2 & 15.8 & 13.6 & 15.0 & 14.2 & 13.7 & 15.2 \\
George Kittle & TE & 10.8 & 13.9 & 13.1 & 13.5 & 13.3 & 13.2 & 0.0 & 14.0 & 14.6 & 14.3 \\
George Pickens & WR & 14.3 & 14.4 & 0.0 & 14.5 & 15.0 & 15.1 & 14.6 & 15.3 & 14.6 & 16.0 \\
Harold Fannin Jr. & TE & 9.6 & 0.0 & 9.6 & 10.4 & 9.6 & 10.1 & 10.1 & 10.3 & 10.1 & 10.1 \\
Harrison Butker & K & 8.8 & 8.5 & 0.0 & 8.3 & 8.7 & 9.1 & 8.3 & 8.7 & 9.0 & 8.4 \\
Hunter Henry & TE & 9.4 & 9.1 & 9.7 & 9.4 & 10.4 & 9.6 & 0.0 & 9.9 & 10.2 & 9.4 \\
Isiah Pacheco & RB & 11.4 & 10.7 & 0.0 & 9.3 & 10.3 & 11.8 & 9.7 & 10.8 & 12.4 & 9.6 \\
J.J. McCarthy & QB & 0.0 & 14.4 & 17.6 & 17.1 & 14.8 & 13.6 & 18.0 & 19.0 & 16.6 & 14.8 \\
J.K. Dobbins & RB & 13.4 & 11.3 & 14.0 & 12.3 & 0.0 & 13.3 & 13.6 & 12.2 & 13.1 & 11.9 \\
Ja'Marr Chase & WR & 19.7 & 23.2 & 0.0 & 23.1 & 23.1 & 23.7 & 22.8 & 23.6 & 22.8 & 21.6 \\
Jacory Croskey-Merritt & RB & 11.5 & 11.2 & 12.2 & 14.1 & 0.0 & 11.4 & 13.5 & 14.9 & 12.6 & 14.9 \\
Jahmyr Gibbs & RB & 0.0 & 19.2 & 18.8 & 17.5 & 20.5 & 17.4 & 20.6 & 15.9 & 19.1 & 18.8 \\
Jake Bates & K & 0.0 & 8.8 & 8.6 & 8.5 & 8.9 & 8.6 & 9.2 & 8.1 & 8.9 & 8.6 \\
Jake Elliott & K & 8.4 & 0.0 & 8.3 & 8.4 & 8.9 & 8.7 & 8.4 & 8.8 & 8.5 & 8.4 \\
Jake Ferguson & TE & 13.1 & 11.5 & 0.0 & 11.5 & 11.9 & 12.0 & 11.6 & 12.3 & 11.6 & 12.9 \\
Jakobi Meyers & WR & 0.0 & 11.5 & 11.5 & 13.2 & 11.1 & 11.7 & 11.5 & 12.0 & 10.7 & 11.8 \\
Jalen Hurts & QB & 21.8 & 0.0 & 21.7 & 21.7 & 26.2 & 24.1 & 22.2 & 23.0 & 24.5 & 23.0 \\
James Cook III & RB & 15.2 & 16.9 & 19.0 & 17.1 & 15.6 & 18.3 & 21.6 & 15.7 & 17.5 & 17.4 \\
Jameson Williams & WR & 0.0 & 11.0 & 11.7 & 11.0 & 10.7 & 10.9 & 11.7 & 10.2 & 11.0 & 11.1 \\
Jauan Jennings & WR & 9.0 & 12.8 & 11.9 & 12.4 & 12.2 & 12.0 & 0.0 & 13.0 & 13.4 & 13.2 \\
Javonte Williams & RB & 16.2 & 16.1 & 0.0 & 16.5 & 15.7 & 15.3 & 14.8 & 17.1 & 16.2 & 16.6 \\
Jaxon Smith-Njigba & WR & 0.0 & 24.4 & 21.7 & 21.3 & 22.4 & 22.8 & 21.0 & 23.7 & 21.2 & 21.4 \\   
Jaxson Dart & QB & 17.8 & 21.1 & 21.1 & 19.5 & 18.6 & 18.9 & 0.0 & 22.3 & 21.6 & 20.2 \\
Jayden Daniels & QB & 0.0 & 19.0 & 20.1 & 21.7 & 0.0 & 19.5 & 22.0 & 22.4 & 21.2 & 24.3 \\
Jayden Reed & WR & 0.0 & 0.0 & 0.0 & 8.7 & 8.9 & 8.5 & 9.0 & 8.5 & 9.0 & 9.1 \\
Jaylen Waddle & WR & 13.1 & 15.1 & 14.5 & 15.4 & 0.0 & 13.3 & 13.8 & 14.5 & 15.3 & 13.9 \\
Jaylen Warren & RB & 16.3 & 15.8 & 16.1 & 19.1 & 17.3 & 16.7 & 16.8 & 17.5 & 15.0 & 15.9 \\
Jaylin Noel & WR & 9.9 & 3.6 & 3.7 & 3.9 & 3.9 & 3.9 & 3.8 & 3.7 & 3.7 & 3.7 \\
Jerry Jeudy & WR & 11.1 & 0.0 & 9.5 & 10.4 & 9.5 & 10.1 & 10.1 & 10.2 & 10.0 & 10.1 \\
Jets D/ST & D/ST & 5.8 & 0.0 & 5.8 & 4.2 & 3.4 & 4.3 & 5.1 & 5.6 & 6.3 & 3.8 \\
Joe Burrow & QB & 0.0 & 0.0 & 0.0 & 0.0 & 0.0 & 0.0 & 0.0 & 0.0 & 0.0 & 0.0 \\
Joe Flacco & QB & 15.4 & 17.3 & 0.0 & 17.1 & 15.8 & 17.4 & 16.1 & 17.8 & 16.2 & 15.6 \\
Joe Mixon & RB & 0.0 & 0.0 & 0.0 & 0.0 & 0.0 & 0.0 & 0.0 & 0.0 & 0.0 & 0.0 \\
Jonathan Taylor & RB & 25.6 & 23.4 & 23.4 & 0.0 & 20.9 & 20.6 & 22.3 & 19.6 & 23.3 & 23.0 \\
Jonnu Smith & TE & 8.1 & 8.7 & 8.2 & 9.0 & 8.6 & 8.5 & 8.8 & 8.4 & 8.1 & 7.8 \\
Jordan Addison & WR & 12.7 & 12.6 & 13.8 & 13.5 & 13.0 & 12.4 & 14.1 & 14.4 & 13.0 & 12.6 \\
Jordan Love & QB & 17.2 & 18.1 & 18.5 & 18.9 & 19.4 & 17.3 & 19.2 & 16.9 & 19.0 & 19.7 \\
Jordan Mason & RB & 11.9 & 6.3 & 7.7 & 8.0 & 6.3 & 5.7 & 7.6 & 8.1 & 8.3 & 6.6 \\
Josh Allen & QB & 20.3 & 20.8 & 21.8 & 20.6 & 18.8 & 22.4 & 24.9 & 20.5 & 19.9 & 21.7 \\
Josh Downs & WR & 10.9 & 11.1 & 10.1 & 0.0 & 11.1 & 9.9 & 10.5 & 10.5 & 11.0 & 10.4 \\
Josh Jacobs & RB & 19.2 & 18.7 & 17.9 & 20.2 & 19.2 & 16.8 & 19.7 & 15.9 & 19.2 & 19.4 \\
Justice Hill & RB & 8.0 & 7.2 & 7.3 & 6.7 & 6.9 & 7.5 & 7.1 & 7.6 & 7.1 & 7.0 \\
Justin Herbert & QB & 20.1 & 21.6 & 21.1 & 19.1 & 0.0 & 20.1 & 19.8 & 18.4 & 22.8 & 17.3 \\
Justin Jefferson & WR & 18.2 & 17.7 & 19.3 & 18.9 & 18.3 & 17.4 & 19.8 & 20.1 & 18.2 & 17.7 \\
Juwan Johnson & TE & 9.6 & 8.7 & 9.0 & 0.0 & 8.8 & 9.6 & 9.4 & 9.0 & 9.2 & 9.7 \\
Kareem Hunt & RB & 7.0 & 6.4 & 0.0 & 5.5 & 6.2 & 7.3 & 5.8 & 6.7 & 7.9 & 5.7 \\
Kayshon Boutte & WR & 7.6 & 9.8 & 10.4 & 10.2 & 11.3 & 10.4 & 0.0 & 10.7 & 11.1 & 10.2 \\
Keenan Allen & WR & 16.0 & 13.9 & 13.9 & 13.1 & 0.0 & 13.2 & 13.7 & 13.7 & 14.9 & 12.3 \\
Kenneth Gainwell & RB & 10.4 & 9.9 & 9.8 & 11.5 & 10.5 & 10.2 & 10.4 & 10.6 & 9.3 & 9.6 \\
Kenneth Walker III & RB & 0.0 & 11.7 & 11.6 & 9.7 & 13.5 & 12.3 & 11.7 & 11.0 & 10.1 & 11.6 \\
Keon Coleman & WR & 10.2 & 9.7 & 9.7 & 9.4 & 8.8 & 9.8 & 10.1 & 9.9 & 9.0 & 9.6 \\
Khalil Shakir & WR & 11.7 & 12.6 & 12.5 & 12.2 & 11.5 & 12.6 & 13.0 & 12.9 & 11.6 & 12.5 \\
Kimani Vidal & RB & 12.7 & 15.5 & 14.0 & 12.9 & 0.0 & 2.9 & 2.6 & 2.5 & 3.0 & 2.4 \\
Kyle Monangai & RB & 8.5 & 9.0 & 8.9 & 8.0 & 8.2 & 7.4 & 7.1 & 7.6 & 7.3 & 7.7 \\
Kyle Pitts Sr. & TE & 10.8 & 10.9 & 10.9 & 9.7 & 9.7 & 9.9 & 9.9 & 10.4 & 10.0 & 9.6 \\
Kyler Murray & QB & 0.0 & 21.4 & 15.9 & 19.3 & 18.2 & 17.1 & 15.8 & 15.5 & 17.8 & 21.7 \\
Kyren Williams & RB & 0.0 & 15.9 & 15.3 & 13.5 & 14.8 & 15.5 & 15.1 & 14.5 & 13.1 & 15.6 \\
Ladd McConkey & WR & 14.8 & 16.0 & 16.1 & 15.2 & 0.0 & 14.6 & 15.2 & 15.2 & 16.5 & 13.7 \\
Lamar Jackson & QB & 0.0 & 23.4 & 24.0 & 21.6 & 24.0 & 26.5 & 24.3 & 26.2 & 22.4 & 22.0 \\
Malik Washington & WR & 9.0 & 10.9 & 10.5 & 11.1 & 0.0 & 9.7 & 10.1 & 10.5 & 11.3 & 10.0 \\
Mark Andrews & TE & 8.6 & 9.9 & 10.0 & 9.2 & 9.5 & 10.4 & 9.9 & 10.5 & 10.0 & 9.7 \\
Marvin Harrison Jr. & WR & 0.0 & 13.1 & 11.2 & 12.2 & 11.5 & 11.7 & 10.9 & 10.6 & 11.1 & 13.0 \\
Matt Gay & K & 0.0 & 7.9 & 8.0 & 8.2 & 0.0 & 7.9 & 8.1 & 8.3 & 8.1 & 8.6 \\
Matthew Golden & WR & 10.3 & 8.3 & 8.7 & 8.5 & 8.7 & 8.2 & 8.7 & 8.2 & 8.7 & 8.9 \\
Matthew Stafford & QB & 0.0 & 18.2 & 19.2 & 17.0 & 17.9 & 18.2 & 17.8 & 17.7 & 16.7 & 17.4 \\
Matthew Wright & K & 0.0 & 0.0 & 0.0 & 0.0 & 0.0 & 0.0 & 0.0 & 0.0 & 0.0 & 0.0 \\
Michael Badgley & K & 8.9 & 8.8 & 8.7 & 0.0 & 8.4 & 8.4 & 8.7 & 8.4 & 8.9 & 8.8 \\
Michael Carter & RB & 0.0 & 7.5 & 5.8 & 2.3 & 2.2 & 2.1 & 1.9 & 1.9 & 2.3 & 2.6 \\
Michael Mayer & TE & 0.0 & 3.4 & 3.4 & 3.9 & 3.3 & 3.4 & 3.4 & 3.5 & 3.1 & 3.5 \\
Michael Penix Jr. & QB & 0.0 & 14.2 & 15.1 & 14.3 & 13.8 & 14.8 & 13.0 & 13.7 & 13.8 & 12.5 \\
Michael Pittman Jr. & WR & 13.8 & 16.3 & 14.8 & 0.0 & 16.4 & 14.7 & 15.4 & 15.5 & 16.2 & 15.3 \\
Michael Wilson & WR & 0.0 & 7.9 & 6.8 & 7.4 & 7.0 & 7.1 & 6.6 & 6.4 & 6.7 & 7.9 \\
Nick Chubb & RB & 9.3 & 8.5 & 9.7 & 11.4 & 9.9 & 9.0 & 8.8 & 9.8 & 10.3 & 9.5 \\
Nick Folk & K & 8.0 & 0.0 & 7.5 & 7.4 & 7.8 & 7.7 & 7.8 & 7.6 & 7.6 & 7.5 \\
Nico Collins & WR & 0.0 & 16.5 & 16.5 & 17.4 & 17.4 & 17.9 & 17.2 & 16.5 & 16.6 & 16.8 \\
Ollie Gordon II & RB & 4.6 & 3.7 & 3.6 & 3.6 & 0.0 & 3.6 & 3.8 & 3.6 & 4.2 & 3.3 \\
Omarion Hampton & RB & 0.0 & 0.0 & 0.0 & 0.0 & 0.0 & 18.4 & 17.1 & 16.4 & 19.4 & 15.8 \\
Oronde Gadsden II & TE & 11.1 & 11.3 & 11.3 & 10.7 & 0.0 & 10.8 & 11.1 & 11.2 & 12.0 & 10.1 \\
Patrick Mahomes & QB & 24.5 & 23.2 & 0.0 & 21.0 & 23.6 & 26.2 & 20.4 & 22.8 & 24.5 & 21.3 \\
Patriots D/ST & D/ST & 5.7 & 4.8 & 4.8 & 7.2 & 5.7 & 6.4 & 0.0 & 4.1 & 4.6 & 7.5 \\
Puka Nacua & WR & 0.0 & 22.2 & 24.2 & 22.6 & 23.3 & 22.5 & 22.9 & 23.0 & 22.7 & 22.1 \\
Quentin Johnston & WR & 13.5 & 12.3 & 12.3 & 11.6 & 0.0 & 11.1 & 11.4 & 11.4 & 12.4 & 10.3 \\
Quinshon Judkins & RB & 12.4 & 0.0 & 15.0 & 14.6 & 14.1 & 13.7 & 17.1 & 14.5 & 13.7 & 14.4 \\
RJ Harvey & RB & 7.2 & 6.7 & 7.6 & 7.3 & 0.0 & 7.8 & 7.5 & 7.2 & 7.3 & 7.2 \\
Rachaad White & RB & 16.9 & 0.0 & 7.5 & 7.9 & 6.9 & 7.9 & 8.0 & 8.0 & 7.9 & 8.3 \\
Raiders D/ST & D/ST & 0.0 & 5.7 & 3.2 & 3.6 & 6.0 & 4.8 & 3.4 & 5.0 & 5.1 & 5.8 \\
Rashee Rice & WR & 17.2 & 16.9 & 0.0 & 16.2 & 17.3 & 18.0 & 15.2 & 16.3 & 16.8 & 16.1 \\
Rashid Shaheed & WR & 12.2 & 11.4 & 11.8 & 0.0 & 11.6 & 12.7 & 12.2 & 11.9 & 12.2 & 12.9 \\
Ray Davis & RB & 1.9 & 2.5 & 2.7 & 2.5 & 2.3 & 2.6 & 2.9 & 2.4 & 2.5 & 2.5 \\
Rhamondre Stevenson & RB & 12.4 & 13.2 & 11.9 & 14.0 & 14.9 & 14.7 & 0.0 & 13.0 & 13.2 & 13.7 \\
Ricky Pearsall & WR & 0.0 & 9.7 & 9.0 & 9.4 & 9.3 & 9.1 & 0.0 & 10.6 & 10.9 & 10.7 \\
Rico Dowdle & RB & 12.7 & 10.8 & 12.2 & 12.0 & 11.7 & 10.4 & 0.0 & 11.9 & 11.4 & 10.2 \\
Rome Odunze & WR & 13.2 & 14.6 & 13.4 & 14.1 & 13.9 & 13.8 & 13.8 & 12.7 & 13.7 & 14.0 \\
Romeo Doubs & WR & 12.2 & 11.6 & 12.3 & 10.9 & 11.3 & 10.7 & 11.3 & 10.8 & 11.3 & 11.5 \\
Ryan Fitzgerald & K & 8.1 & 7.4 & 7.7 & 7.5 & 7.6 & 7.2 & 0.0 & 7.6 & 7.4 & 7.4 \\
Sam Darnold & QB & 0.0 & 17.9 & 15.8 & 14.1 & 17.4 & 17.4 & 15.2 & 16.9 & 14.3 & 15.8 \\
Sam LaPorta & TE & 0.0 & 11.8 & 12.5 & 11.8 & 11.4 & 11.7 & 12.5 & 11.0 & 11.8 & 11.9 \\
Saquon Barkley & RB & 16.9 & 0.0 & 15.3 & 15.6 & 18.3 & 18.0 & 16.3 & 17.4 & 17.1 & 16.5 \\
Steelers D/ST & D/ST & 3.6 & 3.4 & 4.5 & 5.8 & 4.3 & 3.6 & 3.7 & 6.1 & 3.7 & 6.8 \\
Stefon Diggs & WR & 12.8 & 13.1 & 13.9 & 13.6 & 15.1 & 14.0 & 0.0 & 14.4 & 14.8 & 13.6 \\
T.J. Hockenson & TE & 10.2 & 9.5 & 10.5 & 10.3 & 9.9 & 9.4 & 10.8 & 10.9 & 9.8 & 9.5 \\
Tee Higgins & WR & 12.4 & 12.9 & 0.0 & 12.8 & 12.8 & 13.1 & 12.6 & 13.1 & 12.6 & 12.0 \\
Terry McLaurin & WR & 11.8 & 11.6 & 11.8 & 12.3 & 0.0 & 11.8 & 12.5 & 12.1 & 12.2 & 13.2 \\
Tetairoa McMillan & WR & 13.4 & 12.9 & 12.0 & 11.9 & 13.2 & 11.7 & 0.0 & 12.0 & 12.6 & 12.1 \\
Texans D/ST & D/ST & 6.8 & 5.4 & 7.0 & 7.7 & 4.8 & 4.6 & 4.8 & 7.6 & 8.7 & 6.6 \\
Tez Johnson & WR & 8.9 & 0.0 & 5.6 & 5.5 & 5.0 & 5.2 & 5.1 & 5.1 & 5.2 & 5.5 \\
Theo Johnson & TE & 9.5 & 10.1 & 10.2 & 9.9 & 9.6 & 10.3 & 0.0 & 10.6 & 10.2 & 9.5 \\
Tony Pollard & RB & 9.7 & 9.8 & 0.0 & 8.5 & 8.2 & 9.6 & 9.4 & 9.6 & 9.1 & 10.0 \\
Travis Etienne Jr. & RB & 0.0 & 13.1 & 10.9 & 12.7 & 12.4 & 14.8 & 12.2 & 14.0 & 10.9 & 11.8 \\
Travis Hunter & WR & 0.0 & 11.8 & 10.8 & 11.8 & 11.6 & 12.5 & 12.6 & 11.7 & 11.6 & 12.6 \\
Travis Kelce & TE & 10.9 & 10.7 & 0.0 & 10.2 & 10.9 & 11.4 & 9.6 & 10.2 & 10.7 & 10.2 \\
Tre Tucker & WR & 0.0 & 9.4 & 9.3 & 10.8 & 9.1 & 9.5 & 9.3 & 9.7 & 8.7 & 9.8 \\
TreVeyon Henderson & RB & 8.3 & 8.5 & 7.8 & 9.0 & 9.6 & 9.5 & 0.0 & 8.5 & 8.5 & 8.8 \\
Trevor Lawrence & QB & 0.0 & 16.9 & 14.0 & 16.7 & 16.0 & 18.7 & 17.5 & 17.8 & 14.9 & 17.1 \\
Trey Benson & RB & 0.0 & 0.0 & 0.0 & 14.0 & 13.7 & 12.9 & 12.1 & 11.9 & 14.1 & 15.9 \\
Trey McBride & TE & 0.0 & 18.7 & 16.4 & 17.5 & 16.6 & 17.0 & 15.8 & 15.6 & 16.0 & 18.5 \\
Troy Franklin & WR & 10.9 & 10.6 & 11.3 & 11.7 & 0.0 & 12.6 & 11.3 & 11.7 & 11.3 & 11.7 \\
Tucker Kraft & TE & 10.4 & 11.3 & 12.1 & 11.9 & 12.2 & 11.8 & 12.3 & 11.8 & 12.4 & 12.5 \\
Tyjae Spears & RB & 9.5 & 9.5 & 0.0 & 8.4 & 8.3 & 9.4 & 9.1 & 9.5 & 9.1 & 9.7 \\
Tyler Allgeier & RB & 8.0 & 5.4 & 5.9 & 6.7 & 6.4 & 7.0 & 5.3 & 5.7 & 6.2 & 5.4 \\
Tyler Loop & K & 7.9 & 8.5 & 8.5 & 8.2 & 8.9 & 9.1 & 8.8 & 9.0 & 8.3 & 8.3 \\
Tyler Warren & TE & 13.0 & 13.6 & 12.6 & 0.0 & 13.6 & 12.4 & 13.0 & 12.9 & 13.5 & 12.9 \\
Vikings D/ST & D/ST & 4.5 & 3.2 & 3.5 & 3.9 & 2.6 & 4.1 & 3.8 & 2.9 & 5.2 & 3.2 \\
Wan'Dale Robinson & WR & 13.2 & 13.1 & 13.3 & 12.9 & 12.4 & 13.4 & 0.0 & 13.9 & 13.2 & 12.4 \\
Wil Lutz & K & 8.5 & 7.9 & 8.6 & 8.2 & 0.0 & 8.4 & 8.5 & 8.3 & 8.5 & 8.1 \\
Woody Marks & RB & 10.8 & 9.8 & 10.8 & 10.9 & 11.1 & 10.4 & 10.2 & 10.9 & 11.3 & 10.7 \\
Xavier Legette & WR & 8.5 & 6.3 & 6.0 & 5.9 & 6.6 & 5.7 & 0.0 & 6.0 & 6.2 & 6.0 \\
Xavier Worthy & WR & 12.5 & 12.1 & 0.0 & 11.5 & 12.4 & 13.1 & 10.9 & 11.8 & 12.5 & 11.5 \\
Zach Charbonnet & RB & 0.0 & 10.9 & 10.7 & 8.9 & 12.3 & 11.3 & 10.6 & 10.3 & 9.2 & 10.6 \\
Zach Ertz & TE & 9.0 & 8.9 & 9.0 & 9.4 & 0.0 & 9.0 & 9.7 & 9.2 & 9.3 & 10.0 \\
Zay Flowers & WR & 12.8 & 16.1 & 16.3 & 14.9 & 15.4 & 17.1 & 16.3 & 17.1 & 16.4 & 16.0 \\

\end{longtable}



\appendix

\end{document}